\documentclass[10pt,journal,compsoc]{IEEEtran}

\usepackage[cmex10]{amsmath}
\usepackage{acronym}
\usepackage{algorithmic}
\usepackage{array}
\usepackage{mdwmath}
\usepackage{mdwtab}
\usepackage{eqparbox}
\usepackage{fixltx2e}
\usepackage{stfloats}
\usepackage{url}
\usepackage{graphicx}
\usepackage[font=footnotesize,labelfont=bf]{subcaption}
\usepackage{hyperref}

\usepackage[
backend=biber,
style=ieee,
]{biblatex}
\begin{document}
%
\title{Deep learning framework for robot for person detection and tracking}

\author{Adarsh Ghimire,
        Xiaoxiong Zhang,
        Naoufel Werghi,
        Sajid Javed,
        Jorge Dias}

\markboth{Journal of \LaTeX\ Class Files,~Vol.~6, No.~1, January~2007}%
{Shell \MakeLowercase{\textit{et al.}}: Bare Advanced Demo of IEEEtran.cls for Journals}

\vspace{-0.5cm}
\IEEEcompsoctitleabstractindextext{%
\begin{abstract}
Robustly tracking a person of interest in the crowd with a robotic platform is one of the cornerstones of human-robot interaction. The robot platform which is limited by the computational power, rapid movements and occlusions of target requires an efficient and robust framework to perform tracking. This paper proposes a deep learning framework for tracking a person using a mobile robot with stereo camera. The proposed system detects a person based on its head, then utilizes the low cost, high speed regression network based tracker to track the person of interest in real time. The visual servoing of the mobile robot has been designed using PID controller which utilizes tracker output and depth estimation of the person in subsequent frames, hence providing smooth and adaptive movement of the robot based on target movement. The proposed system has been tested in real environment, thus proving its effectiveness.
\end{abstract}

}
\vspace{-1cm}

\maketitle
\IEEEdisplaynotcompsoctitleabstractindextext
\vspace{-1cm}

\section{Introduction}
\IEEEPARstart{T}{raditional}  non-mobile intelligent monitoring and surveillance devices are limited by their stationary nature. For instance, a CCTV camera at airports developed to track specific person, is limited by its field of view. In order to cover up for limited field of vision, some solutions  employed intelligent web of cameras to keep on tracking the subject when it moves. However, these solutions require huge computational  cost and are still limited by its immobile nature when the subject exits the view of the system.

To counter loosing a subject from its view, many researches have taken place by combining robotics and computer vision. This enables a robot with camera to move along with the subject and provide a complete surveillance over it \cite{Integrating_Stereo_Vision_2017, Real-time_3D_2017, Real-time_and_fast_RGB-D_2016, folo_a_vision_2018, Convolutional_Channel_Features-Based_2019}. This type of autonomous robot is a very plausible solution for surveillance task. In addition, it is also widely applicable in health care, entertainment, and industries \cite{A_robust_tracking_algorithm_2021}. For example, such robots can assist handicapped people with household works, help warehouse workers during warehousing, serve nurses in hospitals by carrying medicines and equipments, etc. The major challenges faced by these robotic systems while following a person in crowded environments have been real time and robust tracking requirements.

A person following robot requires knowledge from multiple fields such as person detection, tracking movement, and robot control system. Most works in the literature employ  monocular camera with other sensors to obtain the depth information of the target. \cite{Convolutional_Channel_Features-Based_2019} developed a system that uses an online boosting algorithm with convolutional channel features on  monocular camera feed and depth information from laser range finder sensor to follow the person. \cite{Real-time_3D_2017} used ultrasonic sensor to obtain the depth information of the target and used extended kalman filter in the tracking. Some of the recent works used a stereo vision for simultaneous tracking and depth estimation. \cite{Real-time_and_fast_RGB-D_2016} used HOG-based classifier on stereo feed and unscented kalman filter to track the person while \cite{Integrating_Stereo_Vision_2017} proposed CNN tracker for simultaneously  estimating the target and tracking the person. In addition, \cite{folo_a_vision_2018} utilized SVM and optical flow for tracking the target, while \cite{ A_robust_tracking_algorithm_2021} proposed SVM with HOG features, block matching algorithm for frames, and kalman filter.

In this paper, we present a complete deep learning framework that efficiently follows a person in the crowd in real time. The system uses a head detector for detecting person in the crowd, and then uses high speed regression network tracker\cite{Re3_2018} to track the subject at 150 FPS by incorporating temporal information. The robotic motion has been designed using two PID controllers based on tracker output and depth value. Kobuki robot equipped with a stereo camera is used for experiments.

The main contributions of this paper are : 
\begin{enumerate}
    \item Two wheel driven robot visual servoing algorithm
    \item Appropriate depth estimation from stereo camera
    \item A complete robot system that can track a person in real time
\end{enumerate}
\vspace{-0.4cm}

\section{Method}
\label{method}
\begin{figure}[!h]
        \centering
        \includegraphics[width=\columnwidth, height=0.25\columnwidth]{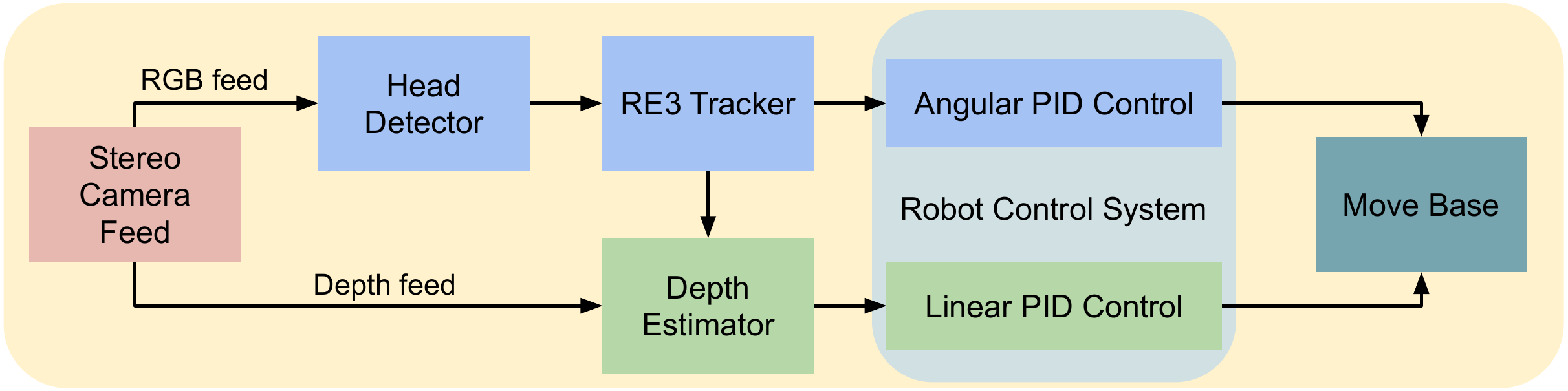}
        \caption{System Block Diagram}
        \vspace{-3em}
        \label{fig:block_digram}
        \end{figure}

Figure \ref{fig:block_digram} shows the block diagram of the proposed person tracking system. Detailed explanation of the complete system is described in sub-sections 2.1, 2.2, 2.3, and, 2.4.\\

\textbf{\hspace{-0.5cm} 2.1 \hspace{0.2cm} Head Detector: }  
The head is the most distinctive  part of a person. Thus, a fast head detector model \cite{Gender_Recognition_2020} has been used to detect people in the feed. The detector outputs bounding boxes and the corresponding probabilities of heads in the first frame. Among the several heads, the algorithm selects the one with the highest probability and sends that to the tracker.
\\
\textbf{\hspace{-0.25cm} 2.2 \hspace{0.2cm} RE3 Tracker: }
The tracker \cite{Re3_2018} first initializes itself with features inside the bounding box given by the detector, then tracks and updates features in subsequent frames. The tracker outputs bounding box of the object in every frame, which is used by the robot control system to generate corresponding motion control signals.
\\
\textbf{\hspace{-0.25cm} 2.3 \hspace{0.2cm} Depth Estimator: } Initial depth value of the object in the current frame is estimated using the median of depth values in the bounding box. We do this to overcome high variance of depth values given by stereo camera. In addition, fast changes in object movements result in transient abrupt robot movements. To address this issue, the final depth value is adjusted according to previous estimates by using exponential weighted moving average.
\\
\textbf{\hspace{-0.25cm} 2.4 \hspace{0.2cm} Robot Control System: }To smooth the movement of the robot, two PID controllers have been designed. First PID controller controls the angular movement of the robot based on the person's horizontal movement on the camera frame tracked by the tracker. Second PID controller controls the linear movement of the robot based on the person's movement towards or away from the robot which is estimated by depth estimator.
\vspace{-0.35cm}
\section{Results}
\begin{figure}[h]
            \begin{subfigure}[c]{0.43\columnwidth}
                \centering
                \includegraphics[width=0.98\columnwidth, height=0.98\columnwidth]{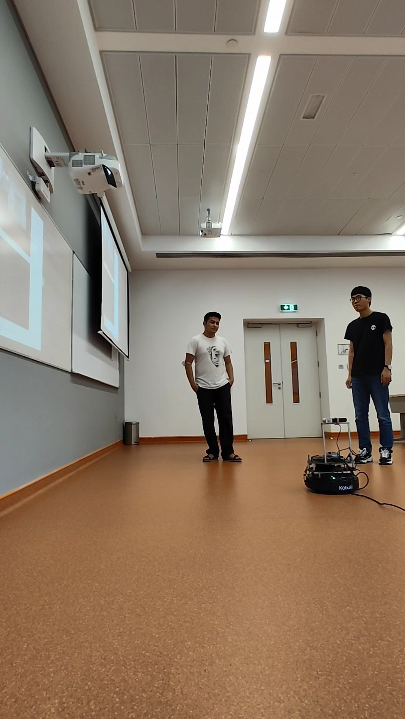}
                \caption{Robot following the person A}
                \label{fig:frame_1}
            \end{subfigure}
            \hspace*{\fill}
            \begin{subfigure}[c]{0.43\columnwidth}
                \centering
                 \includegraphics[width=0.98\columnwidth, height=0.98\columnwidth]{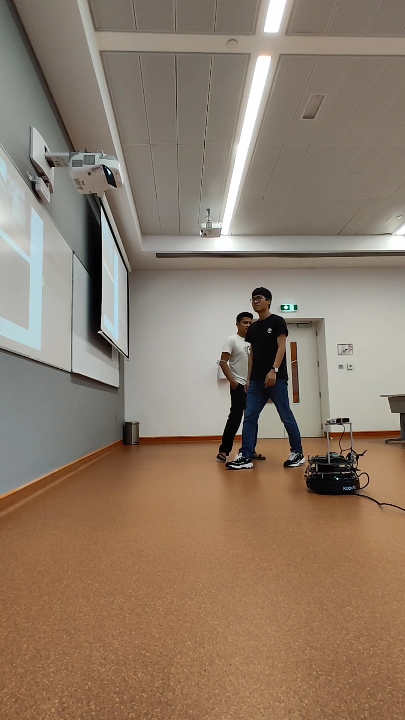}
                \caption{Person A is partially occluded}
                \label{fig:frame_2}
            \end{subfigure}\\
            \begin{subfigure}[c]{0.43\columnwidth}
                \centering
                \includegraphics[width=0.98\columnwidth, height=0.98\columnwidth]{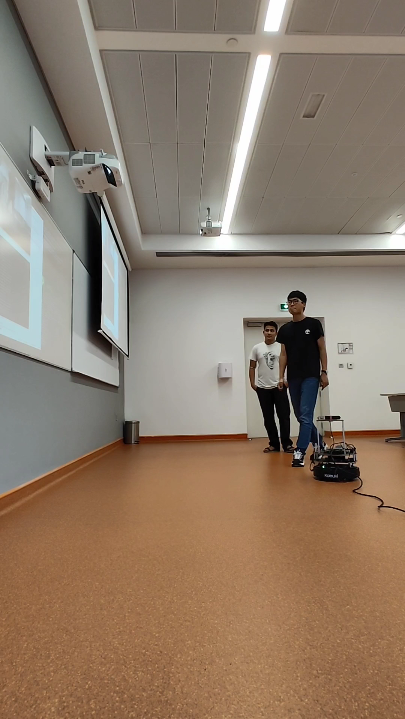}
                \caption{Person A is fully occluded}
                \label{fig:frame_3}
            \end{subfigure}
            \hspace*{\fill}
            \begin{subfigure}[c]{0.43\columnwidth}
                \centering
                \includegraphics[width=0.98\columnwidth, height=0.98\columnwidth]{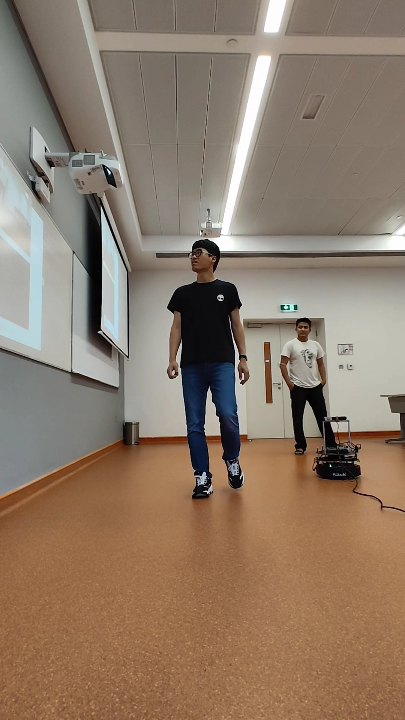}
                \caption{Robot following after occlusion}
                \label{fig:frame_4}
            \end{subfigure}
        \caption{Person Following Robot in the crowd of two people} 
        \vspace{-2.5em}
        \label{fig:results}
        \end{figure}

 Figure \ref{fig:results} reports examples of four different tracking scenarios  where a  robot is following the person A (white dressed) in the presence of person B (black dressed). Full demo can be seen in this link \footnote{\href{https://drive.google.com/file/d/1pekSO87cF3KMdPCyBIlxxj5uDEWeAfvx/view?usp=sharing}{\textit{Video Link}}}.
 
\vspace{-0.4cm}
\section{Conclusion}
In this paper, fast and efficient person following robot system using a real-time recurrent regression network tracker in the context of robotics has been described. The proposed system could perform very well in crowded indoor and outdoor environments. Possible future work includes incorporating more robust and efficient tracker, and more advanced control for the robot.

\vspace{-0.4cm}

\section*{Acknowledgment}
This work acknowledges the support provided by the Khalifa University of Science and Technology under award No. RC1-2018-KUCARS.

\printbibliography
\end{document}